\documentclass[letterpaper]{article} 
\usepackage{aaai25}  
\usepackage{times}  
\usepackage{helvet}  
\usepackage{courier}  
\usepackage[hyphens]{url}  
\usepackage{graphicx} 
\usepackage{natbib}  
\usepackage{caption} 
\usepackage{amssymb}
\usepackage{amsmath}%
\usepackage{bm}%
\usepackage{subcaption}%
\usepackage{xcolor}
\usepackage{dblfloatfix}
\usepackage{appendix}
\usepackage{multirow}
\usepackage{booktabs}
\usepackage{array}
\usepackage{booktabs}
\usepackage{multicol}
\usepackage{ragged2e}
\usepackage[justification=justified]{caption}
\usepackage{enumitem}
\usepackage{algorithm}
\usepackage{algpseudocode}
\usepackage{newfloat}
\usepackage{listings}
\DeclareCaptionStyle{ruled}{labelfont=normalfont,labelsep=colon,strut=off} 
\floatstyle{ruled}
\newfloat{listing}{tb}{lst}{}
\floatname{listing}{Listing}
\title{D$^2$-DPM: Dual Denoising for Quantized Diffusion Probabilistic Models}
\author{
    Qian Zeng\textsuperscript{\rm 1}, 
    Jie Song\textsuperscript{\rm 1}\footnote{Corresponding author.}, 
    Han Zheng\textsuperscript{\rm 1}, 
    Hao Jiang\textsuperscript{\rm 2}, 
    Mingli Song\textsuperscript{\rm 1}
}
\affiliations{
    \textsuperscript{\rm 1} Zhejiang University\\
    \textsuperscript{\rm 2} Alibaba Group

    \{qianz, sjie, h.zheng, haofeizhang, brooksong\}@zju.edu.cn, aoshu.jh@alibaba-inc.com
}

\usepackage{bibentry}

\begin{document}

\maketitle

\begin{abstract}
Diffusion models have achieved cutting-edge performance in image generation. However, their lengthy denoising process and computationally intensive score estimation network impede their scalability in low-latency and resource-constrained scenarios. Post-training quantization (PTQ) compresses and accelerates diffusion models without retraining, but it inevitably introduces additional quantization noise, resulting in mean and variance deviations. In this work, we propose D2-DPM, a dual denoising mechanism aimed at precisely mitigating the adverse effects of quantization noise on the noise estimation network. 
Specifically, we first unravel the impact of quantization noise on the sampling equation into two components: the mean deviation and the variance deviation. The mean deviation alters the drift coefficient of the sampling equation, influencing the trajectory trend, while the variance deviation magnifies the diffusion coefficient, impacting the convergence of the sampling trajectory. The proposed D2-DPM is thus devised to denoise the quantization noise at each time step, and then denoise the noisy sample through the inverse diffusion iterations. 
Experimental results demonstrate that D2-DPM achieves superior generation quality, yielding a 1.42 lower FID than the full-precision model while achieving 3.99x compression and 11.67x bit-operation acceleration. 
\end{abstract}
\begin{links}
\link{Code} {https://github.com/TaylorJocelyn/D2-DPM}
\end{links}

\section{Introduction}\label{sec1}
Diffusion models \cite{sohl2015deep, ho2020denoising, song2019generative, song2020score} have rapidly emerged as predominant deep generative models. By leveraging intricate posterior probability modeling and stable training regimes, diffusion models effectively prevent mode collapse while achieving superior generation fidelity and diversity over GANs \cite{aggarwal2021generative} and VAEs \cite{kingma2013auto}. Recent multi-domain studies demonstrate that highly flexible diffusion models excel in various applications, including text-to-image \cite{Zhu_2023_CVPR}, image super-resolution \cite{wang2023sinsr}, inpainting \cite{lugmayr2022repaint}, style transfer \cite{zhang2023inversion}, text-to-video \cite{singer2022make} and interpretability modeling \cite{lee2022proteinsgm}.

\captionsetup[subfigure]{justification=centering}
\captionsetup{justification=centering, singlelinecheck=false} 
\begin{figure}[t]
    \centering
    \hspace*{-7pt} 
    \includegraphics[width=0.492\textwidth]{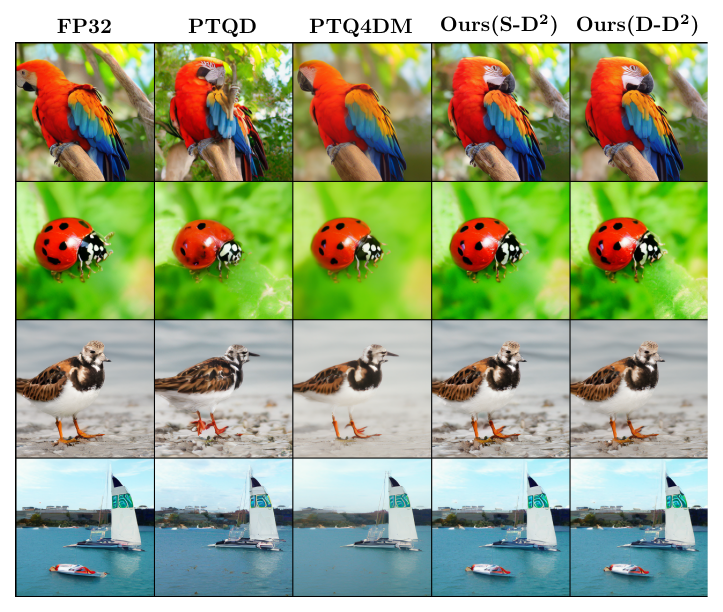}
    \hspace*{\fill} 
    \caption{\justifying Comparison of generated samples on the ImageNet 256$\times$256 between full-precision LDM-4 and its quantized versions using PTQ4DM, PTQD, and our proposed D$^2$-DPM (comprising two variants, S-D$^2$ and D-D$^2$).}
    \label{figure:1}
\end{figure}


However, the generation speed of diffusion models is constrained by two orthogonal factors: the lengthy denoising chain, involving up to 1000 steps \cite{ho2020denoising}, and expensive overhead at each iteration of the cumbersome noise estimation network, \textit{i.e.,} the score estimation network \cite{song2020score}. The former challenge has been significantly alleviated through advanced learning-free samplers, which find more efficient sampling trajectories by providing high-precision numerical approximations for the stochastic differential equations (SDEs) \cite{dockhorn2021score} and ordinary differential equations (ODEs) \cite{lu2022dpm, liu2022pseudo} corresponding to the reverse diffusion process. However, the latter challenge remains formidable. While researchers have employed lightweight model paradigms such as pruning \cite{fang2024structural}, knowledge distillation \cite{ meng2023distillation}, and model quantization \cite{shang2023post, li2023q} to further accelerate each iteration and reduce runtime computational memory overhead for deployment on edge devices, the iterative nature of diffusion models inherently accumulates varying degrees of distortion. 

Specifically, post-training quantization (PTQ) converts a pre-trained FP32 network directly into fixed-point networks with lower bit-width representations for weights and activations, bypassing the necessity for the original training pipeline. It has risen as a commonly embraced methodology owing to its practicality and ease of implementation.  Nonetheless, PTQ unavoidably incurs quantization noise  through the quantification of the noise estimation network, leading to deviations in the estimated mean during the inverse diffusion process and discrepancies with the predetermined variance schedule. The deviation in mean and variance significantly change the sampling trajectory, resulting in a decline in the fidelity of generated images. 

In this work, we initially delve into the pattern of quantization noise based on its statistical properties, based on which we derive the denoising mechanism through the lens of the reverse-time SDE framework \cite{song2020score}. The reparameterization unveils that the mean and standard deviation induced by quantization noise  impact distinct elements of the sampling equation at each time step. The mean deviation is integrated  into the drift term of the reverse-time SDE, altering the direction of the sampling trajectory, while the variance deviation is superimposed on the stochastic term, resulting in increased volatility and divergence of the sampling trajectory. Building upon this insight, we propose \textbf{D$^2$-DPM}, a dual denoising mechanism to precisely mitigate the adverse effects of quantization noise on the noise estimation network. Specifically, we propose to establish a joint Gaussian model for the quantized output and quantization noise at each timestep, enabling precise quantization noise modeling based on the quantized output during inference. With the quantization modeling, we propose two variants of D$^2$-DPM, named \textit{stochastic} dual denoising~(S-D$^{2}$) and \textit{deterministic} dual denoising (D-D$^{2}$), to eliminate the quantization noise during the inverse diffusion process. Experimental results demonstrate that the proposed method achieves significantly superior image generation quality (some examples shown in Fig.~\ref{figure:1}), in both conditional and unconditional image generation tasks.

In conclusion, we summarize our contributions as follow:
\begin{itemize}
\item We make the empirical observation that the quantization noise approximately follow a Gaussian distribution at each time step, which enables precise quantization noise modeling based on the quantized output at inference.
\item  We propose {D$^2$-DPM}, a dual denoising mechanism to precisely mitigate the adverse effects of quantization noise on the noise estimation network. Innovatively, we customize various error correctors for sampling equations with different stochasticity capacities to fully utilize the additional standard deviation.

\item Our extensive experiments demonstrate that, regardless of the stochasticity capacity of the sampling equations, D$^2$-DPM achieves state-of-the-art post-training quantization performance for diffusion models.
\end{itemize}

\section{Related Work}\label{sec2}

\noindent\textbf{Diffusion Model Acceleration.}
To reduce inference costs while maintaining generation quality, two orthogonal diffusion model acceleration methods have emerged: i) optimizing model-agnostic sampling processes, and ii) developing more efficient score estimation models. The former includes advanced techniques such as diffusion scheme learning \cite{chung2022come, franzese2023much}, noise scale learning \cite{kingma2021variational, kong2021fast}, and learning-free samplers based on SDE/ODE acceleration techniques. Meanwhile, the latter leverages model compress paradigms tailored to the intrinsic properties of diffusion models. \cite{salimans2022progressive, song2023consistency, berthelot2023tract} employed knowledge distillation, utilizing ODE formulations parallels mapping the prior distribution to the target distribution through efficient paths within the distribution domain. \cite{fang2024structural} applied Diff-Pruning, achieving approximately a 50\% reduction in FLOPs. \cite{he2023efficientdm} proposed a quantization-aware variant of the low-rank adapter (QALoRA) that can be merged with model weights and jointly quantized to low bit-width. Although these methods significantly lower per-iteration inference overhead, they require retraining the network, resulting in substantial additional time and computational costs.

\noindent\textbf{Model Quantization.} Quantization is a widely used technique for memory compression and computational acceleration \cite{liu2021post, fan2024selective, lin2023awq}. It includes quantization-aware training (QAT) \cite{nagel2022overcoming, chu2024make},  and post-training quantization (PTQ) \cite{nagel2021white, yao2022zeroquant, xiao2023smoothquant}. QAT retrains the network to model quantization noise, incurring significant computational overhead. In contrast, PTQ uses a small calibration dataset to quantize network parameters to low-bit fixed-point values. PTQ commonly employs uniform asymmetric quantization, where the parameters include the scale factor $s$, zero-point $z$ and quantization bit-width $b$. A floating-point value $x$ is quantized to a fixed-point value $x_{\text{int}}$ through the preceding parameters:
\begin{equation}
x_{\text{int}} = \text{clamp}\left(\left\lfloor \frac{x}{s} \right\rceil + z, 0, 2^{b}\right),
\label{eq:1}
\end{equation}
where $\lfloor \cdot \rceil$ is the round operation and clamp is a truncation function.
In practice, quantization parameters are often derived by minimizing the MSE between the pre- and post-quantization weight or activation tensors.

\noindent\textbf{PTQ on Diffusion Models.} Until now, only a limited number of studies have delved into post-training quantization for diffusion models. PTQ4DM \cite{shang2023post} introduces a calibration sampling strategy based on normal distribution but restricts its experimentation to low-resolution datasets. Q-diffusion \cite{li2023q} proposes a time step-aware calibration strategy and shortcut-splitting quantization for Unet. PTQD \cite{he2024ptqd} proposes a PTQ error correction method based on the assumption that the quantization noise is linear correlated with the quantized out, a premise that does not always hold true at various time step. In this work, we adopt a joint Gaussian distribution to modeling the quantization noise, which yield significantly superior performance in diverse experimental settings.

\section{Preliminaries}\label{sec3}
\subsection{Differential Equation Background in Diffusion}\label{subsec3}
Diffusion models progressively add isotropic Gaussian noise with a variance schedule $\beta_1,...,\beta_T\in(0, 1)$ to real data $\mathrm{x}_0$ along the forward propagation chain $\{\mathrm{x}_0,...,\mathrm{x}_T\}$, then approximate the posterior probability $p(\mathrm{x}_{t-1}|\mathrm{x}_t)$ by learning the denoising process. During inference, they iteratively denoise the noisy sample along the learned posterior probability chain $q_{\theta}(\mathrm{x}_{i-1}|\mathrm{x}_i)$ to generate images.

\textbf{The forward SDE.} Song et al. \cite{song2020score} extend the discrete-time propagation chain to continuous-time space by stochastic differential equations. In this theoretical framework, the diffusion process can be modeled as a solution to an Itô SDE:
\begin{equation}
\mathrm{d}\mathbf{x} = \mathbf{f}(\mathbf{x}, t) \, \mathrm{d}t + g(t) \, \mathrm{d}\mathbf{w},
\label{eq:2}
\end{equation}
where $\mathbf{w}$ is the standard Wiener process (\textit{a.k.a.}, Brownian motion), $\mathbf{f}(\cdot, t):\mathbb{R}^d \rightarrow \mathbb{R}^d$ is a vector-valued function called the drift coefficient of $\mathbf{x}(t)$, and $g(\cdot):\mathbb{R} \rightarrow \mathbb{R}$ is a scalar function known as the diffusion coefficient of $\mathbf{x}(t)$, representing the stochasticity capacity. 

\textbf{Inverse Sampling Equation.} The predominant sampling methodologies are categorized into \textit{stochastic} and \textit{deterministic} sampling. Stochastic sampling follows Anderson's reverse-time SDE \cite{anderson1982reverse}: 
\begin{equation}
\mathrm{d}\mathbf{x} = \left[ \mathbf{f}(\mathbf{x}, t) - g(t)^2 \nabla_{\mathbf{x}} \log p_t(\mathbf{x}) \right] \mathrm{d}t + g(t) \mathrm{d}\bar{\mathbf{w}},
\label{eq:3}
\end{equation} 
where $\nabla_{\mathbf{x}} \log p_t(\mathbf{x})$ is the score function \cite{bao2022analytic}, $\bar{\mathbf{w}}$ is a standard Wiener process when time flows backwards from T to 0, and dt is an infinitesimal negative timestep.

Deterministic sampling is typically formalized as the Probability Flow ODE sampling equation, derived by \cite{song2020score} from the Fokker-Planck equation, ensuring that the corresponding probability densities $p_t(\mathbf{x})$ of the reverse-time SDE and ODE are equivalent at any given time $t$. The equation is as follows:
\begin{equation}
\mathrm{d}\mathbf{x} = \left[ \mathbf{f}(\mathbf{x}, t) - \frac{1}{2} g(t)^2 \nabla_{\mathbf{x}} \log p_t(\mathbf{x}) \right] \mathrm{d}t,
\label{eq:4}
\end{equation}

To estimate the score $\nabla_{\mathbf{x}} \log p_t(\mathbf{x})$ in Eqn.~\eqref{eq:3} and Eqn.~\eqref{eq:4}, it is common to train a time-independent score-based model $\mathbf{s_\theta}(\mathrm{\mathbf{x}}_t, t)$, which is linearly related to the denoising network:
\begin{equation}
\mathrm{\mathbf{s_\theta}}(\mathbf{x}, t)\triangleq - \frac{\mathrm{\boldsymbol{\epsilon}}_\theta(\mathbf{x}_t, t)}{\sigma_t},
\label{eq:5}
\end{equation} 
where $\sigma_t$ is the standard deviation of $p_{0t}(\mathbf{x}_t|\mathbf{x}_0)$, referred to as the noise schedule and $\mathrm{\boldsymbol{\epsilon}}_\theta(\mathbf{x}_t, t)$ is the noise prediction network, from which quantization noise is introduced.

ODE-based samplers achieve faster sampling speeds due to the deterministic nature of their components. SDE-based samplers leverage the stochasticity provided by stochastic term $g(t)\mathrm{d}\bar{\mathbf{w}}$ to achieve better generation quality. This stochasticity essentially functions as implicit Langevin diffusion, driving the sample towards the desired marginal distribution over time while correcting any errors made in earlier sampling steps. Inspired by this, we utilize the effective components of quantization errors to supplement stochasticity, thereby ensuring high generation quality.
\renewcommand{\figurename}{Figure}
\renewcommand{\figurename}{Figure}
\captionsetup[subfigure]{justification=centering}
\captionsetup{justification=raggedright, singlelinecheck=false} 

\begin{figure*}[h!]
    \centering
    \begin{subfigure}{0.29\textwidth}
        \centering
        \includegraphics[width=\textwidth]{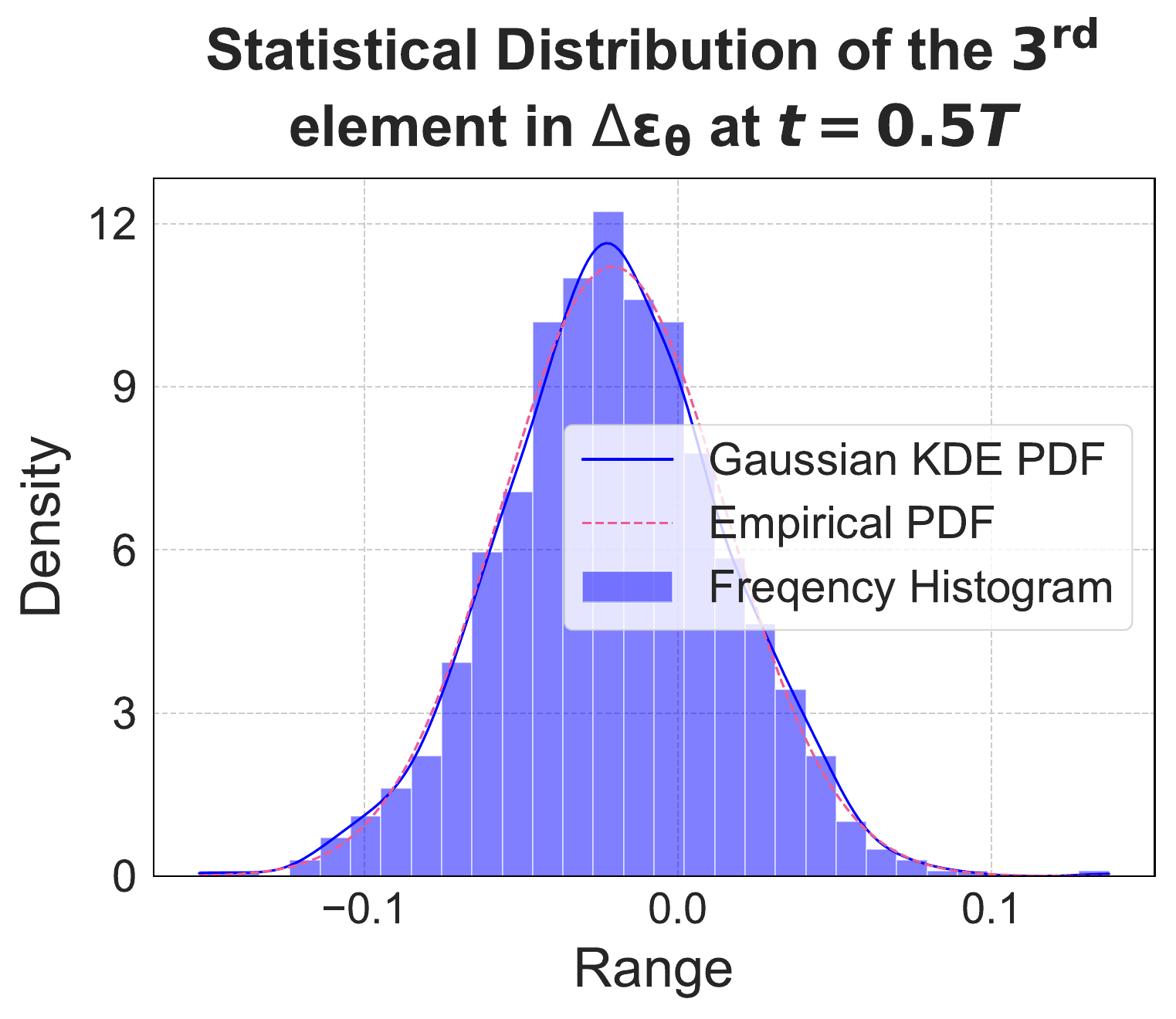}
        \caption{}
        \label{figure:2_a}
    \end{subfigure}
    \hspace{15pt}
    \begin{subfigure}{0.29\textwidth}
        \centering
        \includegraphics[width=\textwidth]{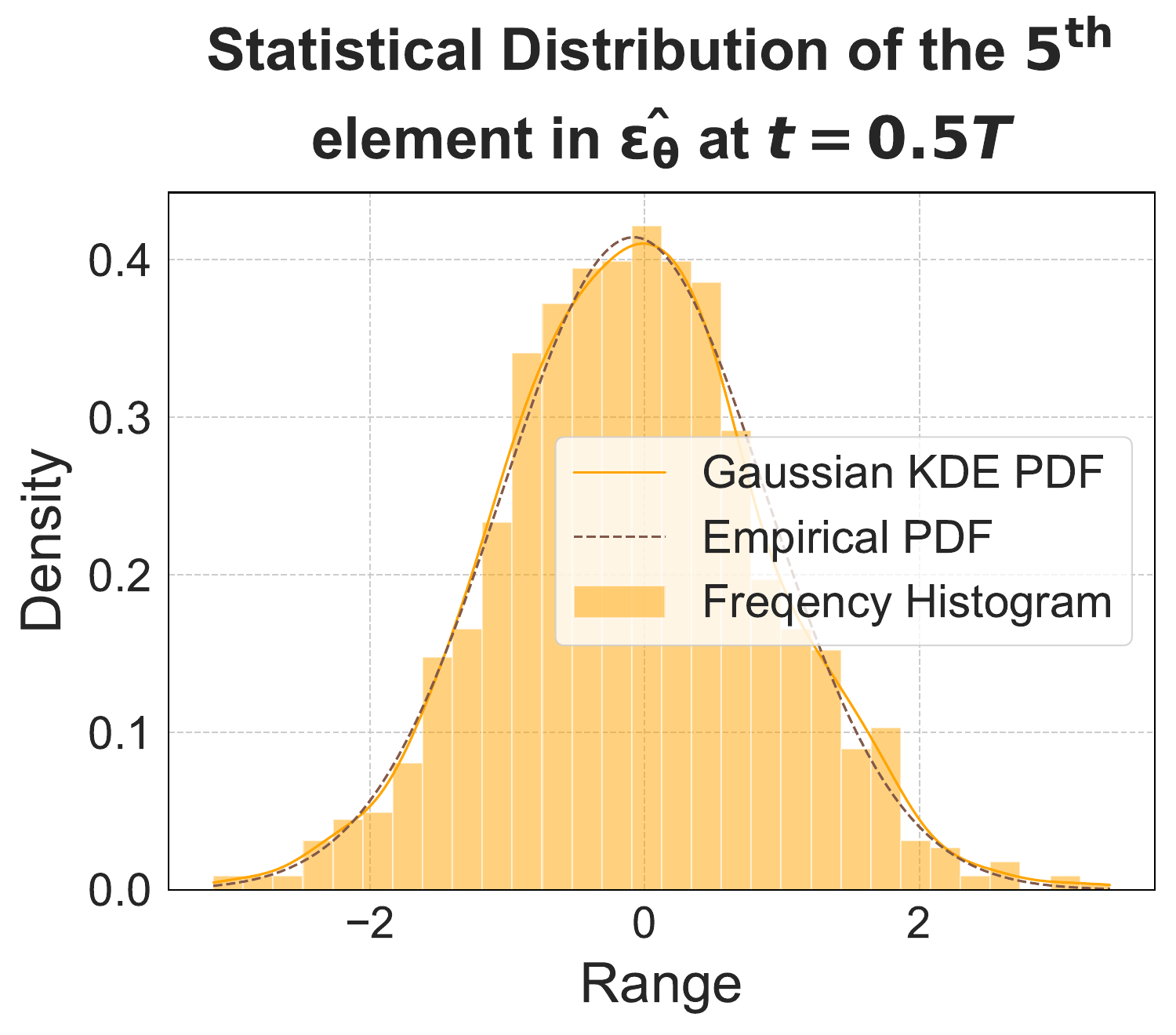}
        \caption{}
        \label{figure:2_b}
    \end{subfigure}
    \hspace{15pt}
    \begin{subfigure}{0.325\textwidth}
        \centering
        \includegraphics[width=\textwidth]{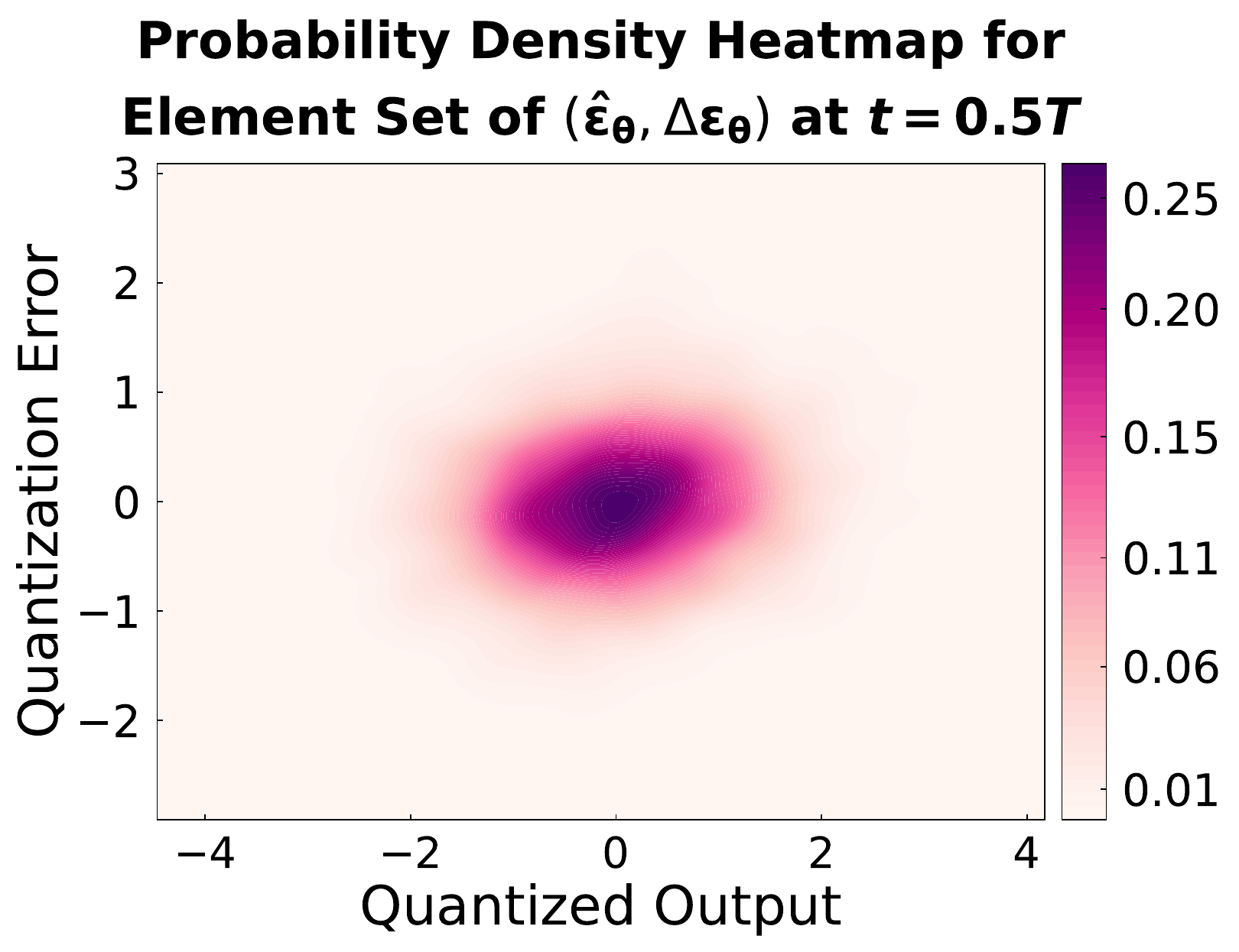}
        \caption{}
        \label{figure:2_c}
    \end{subfigure}
    \hspace{-1pt}

    \caption{\justifying The statistical characteristics of $\Delta\bm{\epsilon}_{\theta}$ and $\bm{\hat{\epsilon}}_{\theta}$ {{on quantifying full-precision LDM-4~\cite{rombach2022high} to W4A8 (4-bit for weights, 8-bit for activations) LDM-4}}. \textbf{(a)} The statistical distribution of the $3^{rd}$ element of $\Delta\bm{\epsilon}_{\theta}^{(0.5T)}$. \textbf{(b)}  The statistical distribution of the $5^{th}$ element of $\bm{\hat{\epsilon}}_{\theta}^{(0.5T)}$. \textbf{(c)}  The probability density heatmap for element set of $\left(\bm{\hat{\epsilon}_{\theta}^{(0.5T)}}, \Delta\bm{\epsilon_{\theta}^{(0.5T)}}\right)$.}
    \label{figure:2}
\end{figure*}

\subsection{Pre-analysis: Quantization Noise on Diffusion}\label{sec:3.2}

We first make an empirical analysis of the quantization noise incurred by the quantized noise estimation model. At time step $t$, we use $\bm{\epsilon}_{\theta}^{(t)}$ to denote the full-precision output, $\bm{\hat{\epsilon}}_{\theta}^{(t)}$ the quantized output, and $\Delta\bm{\epsilon}_{\theta}^{(t)}$ the quantization noise, $\Delta\bm{\epsilon}_{\theta}^{(t)}=\bm{\hat{\epsilon}}_{\theta}^{(t)} - \bm{\epsilon}_{\theta}^{(t)}$. Fig.~\ref{figure:2} depicts some example results on LDM-4 \cite{rombach2022high} with the W4A8 (4-bit weights and 8-bit activations) quantization. Experiments on more diffusion models draw similar conclusions and are provided in the Appendix. From these results, we make following main findings.

\noindent\textbf{Observation \#1}: \textit{At each time step, the quantization noise $\Delta\bm{\epsilon}_{\theta}^{(t)}$ approximately follows a Gaussian distribution: $\Delta\mathrm{\boldsymbol{\epsilon}}_\theta^{(t)} \sim \mathcal{N}(\mathrm{\bm{\mu_{\Delta}}}(t), \bm{\Sigma_{\Delta}}(t))$.}

As a non-cherrypick example, the distribution of the $3^{rd}$ element in $\Delta\bm{\epsilon}_{\theta}^{(0.5T)}$ is illustrated in the Fig.~\ref{figure:2_a}. More results can be found in the Appendix. It is evident that the KDE-fitted probability density curve and the Gaussian density curve overlap almost exactly, which leads to the above observation.

\noindent\textbf{Observation \#2}: \textit{At each time step, the quantized output $\mathrm{\boldsymbol{\hat{\epsilon}}}_\theta^{(t)}$ approximately follows a Gaussian distribution:  $\mathrm{\boldsymbol{\hat{\epsilon}}}_\theta^{(t)} \sim \mathcal{N}(\mathrm{\bm{\mu_{\hat{\epsilon}}}}(t), \bm{\Sigma_{\hat{\epsilon}}}(t))$. }

This is a straightforward finding as the full-precision output $\bm{\epsilon}_{\theta}^{(t)}$ follows a Gaussian distribution by its nature. If the quantization noise $\Delta\bm{\epsilon}_{\theta}^{(t)}$ follows a Gaussian distribution (\textit{Observation \#1}), then the quantized output $\bm{\hat{\epsilon}}_{\theta}^{(t)}$ also follows a Gaussian distribution. Fig.~\ref{figure:2_b} depicts distribution of the $5^{th}$ element of $\bm{\hat{\epsilon}}_{\theta}^{(0.5T)}$, which showcase that the quantized output $\mathrm{\boldsymbol{\hat{\epsilon}}}_\theta^{(t)}$ approximately follows a Gaussian distribution. 

\subsection{Quantization Noise on SDE}
With quantization noise $\Delta\mathrm{\boldsymbol{\epsilon}}_\theta^{(t)} \sim \mathcal{N}(\mathrm{\bm{\mu_{\Delta}}}(t), \bm{\Sigma_{\Delta}}(t))$ from Observation \#1,  the SDE-based sampling with quantization noise can be reformulated as follows:
\begin{align}
        \mathrm{d}\mathbf{x} &= \left[ \mathbf{f}(\mathbf{x}, t) + g(t)^2 \frac{\mathrm{\boldsymbol{\epsilon}_\theta}(\mathbf{x}_t, t) + \mathrm{\bm{\mu}}_\Delta(t)}{\sigma_t} \right] \mathrm{d}t \nonumber \\
        &+ \left[g(t) + \frac{g(t)^2 \sigma_\Delta(t)\sqrt{\mathrm{d}t}}{\sigma_t}\right]\mathrm{d}\bar{\mathbf{w}},
    \label{eq:7}
\end{align}
where $\sigma_\Delta(t)$ represents the standard deviation of the isotropic standard Gaussian component of $\Sigma_\Delta(t)$. From the above equation, it is evident that $\bm{\mu}_\Delta(t)$ and $\sigma_\Delta(t)$ independently affect the inverse sampling equation. Specifically, the mean alters the drift term, altering the sampling direction, while the variance increases the diffusion coefficient, impacting the fluctuation and convergence of the sampling trajectory. Therefore, we separately formulate equations for expectation and variance to conduct the analysis, followed by performing mean and variance corrections in a fully disentangled manner.

\section{The Proposed Method}\label{sec4}

\subsection{Time Step-aware Quantization Noise Modeling}\label{sec:4.1}

For the sake of clarity, we use $\boldsymbol{\hat{\mathcal{E}}}$ and $\Delta\boldsymbol{ \mathcal{E}}$ to denote the variables of $\bm{\hat{\epsilon}}$ and $\Delta\bm{\epsilon}$, respectively. With Observation \#1 and \#2, we can employ a Gaussian distribution to model the joint distribution of $\boldsymbol{\hat{\mathcal{E}}}$ and $\Delta\boldsymbol{ \mathcal{E}}$ (as shown in Fig.~\ref{figure:2_c}):
\begin{equation}
    \begin{bmatrix}
    \boldsymbol{\hat{\mathcal{E}}} \\
     \Delta\boldsymbol{\mathcal{E}}
    \end{bmatrix} \sim \mathcal{N} \left( \begin{bmatrix}
    \boldsymbol{\hat{\mu}} \\
    \Delta\boldsymbol{\mu}
    \end{bmatrix}, \begin{bmatrix}
    \boldsymbol{\Sigma}_{\boldsymbol{\hat{\mathcal{E}}},\boldsymbol{\hat{\mathcal{E}}}} &  \boldsymbol{\Sigma}_{\boldsymbol{\hat{\mathcal{E}}}, \boldsymbol{\Delta \mathcal{E}}} \\
    \boldsymbol{\Sigma}_{\boldsymbol{\Delta \mathcal{E}},\boldsymbol{\hat{\mathcal{E}}}}
      & \boldsymbol{\Sigma}_{\boldsymbol{\Delta \mathcal{E}},\boldsymbol{\Delta \mathcal{E}}}
    \end{bmatrix} \right)
    \label{eq:joint_modeling}
\end{equation}
Note that at different time step $t$, $\boldsymbol{\hat{\mathcal{E}}}$ and $\Delta\boldsymbol{ \mathcal{E}}$ jointly follow a different Gaussian distribution (thus termed \textit{time step-aware}). Here we omit the time step $t$ for the symbol simplicity. $\boldsymbol{\hat{\mu}}$ and $\Delta\boldsymbol{ \mu}$ denote the mean of the quantized output and the quantization noise, respectively. $\boldsymbol{\Sigma}_{\boldsymbol{\hat{\mathcal{E}}}, \Delta\boldsymbol{ \mathcal{E}}}$ denote the cross-covariance between the quantized output $\boldsymbol{\hat{\mathcal{E}}}$ and the quantization noise $\Delta\boldsymbol{ \mathcal{E}}$. With Eqn.~\ref{eq:joint_modeling}, we can derive the distribution of the quantization noise $\Delta\boldsymbol{ \mathcal{E}}$ conditioned on the quantized output $\boldsymbol{\hat{\mathcal{E}}}=\boldsymbol{\hat{\epsilon}}$ as follows:
\begin{equation}
    \left\{ \Delta\boldsymbol{ \mathcal{E}} | \boldsymbol{\hat{\mathcal{E}}}=\boldsymbol{\hat{\epsilon}} \right\} \sim \mathcal{N}\left( \boldsymbol{\mu}_{\Delta\boldsymbol{ \mathcal{E}} | \boldsymbol{\hat{\mathcal{E}}}=\boldsymbol{\hat{\epsilon}}}, \boldsymbol{\Sigma}_{\Delta\boldsymbol{ \mathcal{E}} | \boldsymbol{\hat{\mathcal{E}}}=\boldsymbol{\hat{\epsilon}}}
    \right),
    \label{eq:condition_distribution}
\end{equation}
\begin{equation}
    \boldsymbol{\mu}_{\Delta\boldsymbol{ \mathcal{E}} | \boldsymbol{\hat{\mathcal{E}}}=\boldsymbol{\hat{\epsilon}}} = \boldsymbol{\Sigma}_{\boldsymbol{\Delta \mathcal{E}},\boldsymbol{\hat{\mathcal{E}}}}
    \boldsymbol{\Sigma}_{\boldsymbol{\hat{\mathcal{E}}},\boldsymbol{\hat{\mathcal{E}}}}^{-1}
    (\boldsymbol{\hat{\epsilon}}-\boldsymbol{\hat{\mu}}) + \Delta\boldsymbol{\mu},
\end{equation}
\begin{equation}
    \boldsymbol{\Sigma}_{\Delta\boldsymbol{ \mathcal{E}} | \boldsymbol{\hat{\mathcal{E}}}=\boldsymbol{\hat{\epsilon}}} = \boldsymbol{\Sigma}_{\boldsymbol{\Delta \mathcal{E}},\boldsymbol{\Delta \mathcal{E}}} - 
    \boldsymbol{\Sigma}_{\boldsymbol{\Delta \mathcal{E}},\boldsymbol{\hat{\mathcal{E}}}}
    \boldsymbol{\Sigma}_{\boldsymbol{\hat{\mathcal{E}}},\boldsymbol{\hat{\mathcal{E}}}}^{-1}
    \boldsymbol{\Sigma}_{\boldsymbol{\hat{\mathcal{E}}},\boldsymbol{\Delta \mathcal{E}}}.
\end{equation}
However, directly estimating the joint distribution in Eqn.~\ref{eq:joint_modeling} can be problematic due to the high dimensions of the joint space of $\boldsymbol{\hat{\mathcal{E}}}$ and $\Delta\boldsymbol{ \mathcal{E}}$. In this work, we make the assumption that elements in $\boldsymbol{\hat{\mathcal{E}}}$ ($\Delta\boldsymbol{ \mathcal{E}}$) are uncorrelated (but the $i$-th element in $\boldsymbol{\hat{\mathcal{E}}}$ can be correlated to the $i$-th element in $\Delta\boldsymbol{ \mathcal{E}}$). With the assumption, the covariance matrices $\boldsymbol{\Sigma}_{\boldsymbol{\hat{\mathcal{E}}},\boldsymbol{\hat{\mathcal{E}}}}$, $\boldsymbol{\Sigma}_{\boldsymbol{\hat{\mathcal{E}}}, \boldsymbol{\Delta \mathcal{E}}}$, $\boldsymbol{\Sigma}_{\boldsymbol{\Delta \mathcal{E}},\boldsymbol{\hat{\mathcal{E}}}}$ and $\boldsymbol{\Sigma}_{\boldsymbol{\Delta \mathcal{E}},\boldsymbol{\Delta \mathcal{E}}}$ become diagonal matrices. We further assume the distributions of $\boldsymbol{\hat{\mathcal{E}}}$ and $\Delta\boldsymbol{ \mathcal{E}}$ to be isotropic (\textit{i.e.,} $\bm{\Sigma}=\sigma^2\bm{I}$), which significantly simplify the estimation of the joint distribution.

\subsection{The Proposed D$^2$-DPM}\label{sec:4.2}
Now we provide the proposed {dual denoising} mechanism. We coin the proposed method {``dual denoising''} as it denoises two types of noise during the inverse diffusion process, including \textit{quantization noise} and \textit{diffusion noise}. Specifically, we propose two variants of dual denoising, named \textit{stochastic} dual denoising~(S-D$^{2}$) and \textit{deterministic} dual denoising~(D-D$^{2}$). 

\subsubsection{Stochastic Dual Denoising.}\label{sec:4.2.1}

In the stochastic variant of dual denoising, we recover the distribution of the diffusion noise by subtracting the estimated quantization noise $\Delta\bm{\mathcal{E}}^{'}$ from the quantized output $\bm{\hat{\mathcal{E}}}$:
\begin{align}
    \bm{\mathcal{E}}^{'} &=  \bm{\hat{\mathcal{E}}} - \Delta\bm{\mathcal{E}}^{'}.
\end{align}
Obviously $\bm{\mathcal{E}}^{'}$ also follows a Gaussian distribution. If the estimated quantization noise $\Delta \bm{\mathcal{E}}^{'}$ accurately captures the real quantization noise $\Delta \bm{\mathcal{E}}$, then the expectation and the covariance matrix of the recovered diffusion noise can be derived as follows:
\begin{align}
    \mathbf{E}[\bm{\mathcal{E}}^{'}] &= \mathbf{E}[\bm{{\mathcal{E}}}] + \mathbf{E}[\Delta\bm{{\mathcal{E}}}] - \mathbf{E}[\mathbf{\Delta\bm{\mathcal{E}}}^{'}]   
    = \mathbf{E}[\bm{{\mathcal{E}}}]
\end{align}
\begin{align}
    \mathbf{Var}[\bm{\mathcal{E}}^{'}] &= \mathbf{Var}[\bm{{\mathcal{E}}}] + \mathbf{Var}[\Delta\bm{{\mathcal{E}}}] + \mathbf{Var}[\mathbf{\Delta\bm{\mathcal{E}}}^{'}] \nonumber \\ 
    & + 2\mathbf{Cov}[\bm{{\mathcal{E}}}, \Delta\bm{{\mathcal{E}}}] - 2\mathbf{Cov}[\bm{{\mathcal{E}}}, \mathbf{\Delta\bm{\mathcal{E}}}^{'}] \nonumber \\
    & - 2\mathbf{Cov}[\Delta\bm{{\mathcal{E}}},\mathbf{\Delta\bm{\mathcal{E}}}^{'}] \nonumber \\
    &=\mathbf{Var}[\bm{{\mathcal{E}}}]
\end{align}
It can be seen that recovered diffusion noise follows the same distribution as that of the original diffusion noise. The sampling can be achieved by solving the following SDE:
\begin{align}
    \mathrm{d}\mathbf{x} = & \left[ \mathbf{f}(\mathbf{x}, t) + g(t)^2 \frac{\mathrm{\hat{\boldsymbol{\epsilon}}}_\theta(\mathbf{x}_t, t) - \Delta\boldsymbol{\epsilon}^{'}(\mathbf{x}_t,t)}{\sigma_t} \right] \mathrm{d}t \nonumber\\
    &+ g(t)\mathrm{d}\bar{\mathbf{w}}
\end{align}

\subsubsection{Deterministic Dual Denoising}
In the deterministic variant of dual denoising, the distribution of the diffusion noise is recovered by subtracting the mean vector  $\Delta\bm{\mu}$ of quantization noise  from the quantized output $\bm{\hat{\mathcal{E}}}$:

\begin{align}
    \bm{\mathcal{E}}^{'} &=  \bm{\hat{\mathcal{E}}} - \Delta\bm{\mu}.
\end{align}
$\bm{\mathcal{E}}^{'}$ again follows a Gaussian distribution. The expectation and the covariance matrix of the recovered diffusion noise can be derived as follows:
\begin{align}
    \mathbf{E}[\bm{\mathcal{E}}^{'}] &= \mathbf{E}[\bm{{\mathcal{E}}}] + \mathbf{E}[\Delta\bm{{\mathcal{E}}}] - \mathbf{E}[\mathbf{\Delta\bm{\mu}}]   
    = \mathbf{E}[\bm{{\mathcal{E}}}]
\end{align}
\begin{align}
    \mathbf{Var}[\bm{\mathcal{E}}^{'}] &= \mathbf{Var}[\bm{{\mathcal{E}}}] + \mathbf{Var}[\Delta\bm{{\mathcal{E}}}] + \mathbf{Var}[\mathbf{\Delta\bm{\mu}}] \nonumber \\ 
    & + 2\mathbf{Cov}[\bm{{\mathcal{E}}}, \Delta\bm{{\mathcal{E}}}] - 2\mathbf{Cov}[\bm{{\mathcal{E}}}, \mathbf{\Delta\bm{\mu}}] \nonumber \\
    & - 2\mathbf{Cov}[\Delta\bm{{\mathcal{E}}},\mathbf{\Delta\bm{\mu}}] \nonumber \\
    &=\mathbf{Var}[\bm{{\mathcal{E}}}] + \sigma^2_{\Delta}\bm{I}
\end{align}
It can be seen that deterministic dual denoising introduces additional variance $\sigma^2_{\Delta}\bm{I}$, which can be absorbed into diffusion term:
\begin{align}
    \mathrm{d}\mathbf{x} = & \left[ \mathbf{f}(\mathbf{x}, t) + g(t)^2 \frac{\mathrm{\hat{\boldsymbol{\epsilon}}}_\theta(\mathbf{x}_t, t) - \Delta\boldsymbol{\mu}}{\sigma_t} \right] \mathrm{d}t \nonumber\\
    &+ \sqrt{g^2(t)-\frac{g^4(t)\sigma_{\Delta}^2(t)}{\sigma^2_{t}}}\mathrm{d}\bar{\mathbf{w}}
\end{align}
Algorithm~\ref{alg:dddpm} summarizes the procedure of the proposed dual denoising mechanism.

\algrenewcommand\algorithmicrequire{\textbf{Input:}}
\algrenewcommand\algorithmicensure{\textbf{Output:}}
\begin{algorithm}[t!]
\caption{The Methodological Framework of D$^2$-DPM}
\textbf{Description:} 
$S$: the number of sampling times; $M$: the number of sampling steps; TSQNM: Time Step-aware Quantization Noise Modeling. $\{\bm{\Sigma}_t = \sigma_t^2\bm{I}\}_{t=1}^T$ and $\{\alpha_t\}_{t=1}^T$ are parameters within the DDIM sampler.

\hrule 
\begin{algorithmic}[1]
\Require{Full precision model $model_{fp}$, quantization parameters $q_{params}$, sampling inputs $\{\mathbf{x}_T^i\}_{i=1}^N \sim \mathcal{N}(0, \bm{I})$}
\Ensure{Generated samples $\{\mathbf{x}_0^i\}_{i=1}^N$}
\State \small $\{(\mathbf{x}_t, t, c)^i\}_{i=1}^{M \times T} = \text{collect\_calibration}(model_{fp})$ 
\State \fontsize{7.8}{16} $model_q = \mathbf{BRECQ}(model_{fp}, q_{params}, \{(\mathbf{x}_t, t, c)^i\}_{i=1}^{M \times T})$

\State \fontsize{7.67}{16} $\{(\boldsymbol{\hat{\mathcal{E}}}, \Delta\boldsymbol{\mathcal{E}})^i\}_{i=1}^{S \times T} = $ \text{collect\_quant\_error}$(model_{fp}, model_q)$

\State \fontsize{9}{16} $\bm{\mu}_{T \times 2}$, $\bm{\Sigma}_{T \times 4} = $ \text{gaussian\_modeling}($\{(\boldsymbol{\hat{\mathcal{E}}}, \Delta\boldsymbol{\mathcal{E}})^i\}_{i=1}^{S \times T}$)
\For{\text{sample\_num} $i$ in $1, \dots, N$}
    \State $\mathbf{x}_T = \mathbf{x}_T^i$
    \For{timestep $t$ in $T, \dots, 1$}
    \State $\bm{\hat{\epsilon}}_{\theta}^{(t)} = model_q(\mathbf{x}_t)$ 
    
        \State \fontsize{8.6}{16} $\boldsymbol{\mu}_{\Delta\boldsymbol{\mathcal{E}} | \boldsymbol{\hat{\mathcal{E}}} = \boldsymbol{\hat{\epsilon}}_{\theta}^{(t)}}, \boldsymbol{\Sigma}_{\Delta\boldsymbol{\mathcal{E}} | \boldsymbol{\hat{\mathcal{E}}} = \boldsymbol{\hat{\epsilon}}_{\theta}^{(t)}}$  
         =
         \text{TSQNM}($\bm{\hat{\epsilon}}_{\theta}^{(t)}, \bm{\mu}[t], \bm{\Sigma}[t]$)

    \If{\textit{stochastic}
dual denoising}
        \State $\mathbf{z} \sim \mathcal{N}(0, \bm{I})$
        \State $\Delta\bm{\mathcal{E}}^{'} = \boldsymbol{\mu}_{\Delta\boldsymbol{\mathcal{E}} | \boldsymbol{\hat{\mathcal{E}}} = \boldsymbol{\hat{\epsilon}}_{\theta}^{(t)}} + \boldsymbol{\Sigma}_{\Delta\boldsymbol{\mathcal{E}} | \boldsymbol{\hat{\mathcal{E}}} = \boldsymbol{\hat{\epsilon}}_{\theta}^{(t)}}^{1/2} \cdot \mathbf{z}$
        \State $\bm{\mathcal{E}}^{'} \gets \bm{\hat{\epsilon}}_{\theta}^{(t)} - \Delta\bm{\mathcal{E}}^{'}$
    \Else
        \State \fontsize{8.6}{20} $\bm{\mathcal{E}}^{'} \gets \bm{\hat{\epsilon}}_{\theta}^{(t)} - \boldsymbol{\mu}_{\Delta\boldsymbol{\mathcal{E}} | \boldsymbol{\hat{\mathcal{E}}} = \boldsymbol{\hat{\epsilon}}_{\theta}^{(t)}}$
        \State \fontsize{8.6}{20} $k = \sqrt{1 - \alpha_{t-1} - \left|\bm{\Sigma}_t \right|^{1/d}}
         - \sqrt{\frac{\alpha_{t-1}\left(1-\alpha_t\right)}{\alpha_t}}$
        \State \fontsize{8.6}{20} $
            \boldsymbol{\Sigma}_t \gets \boldsymbol{\Sigma}_t - k^2 \cdot \boldsymbol{\Sigma}_{\Delta\boldsymbol{\mathcal{E}} | \boldsymbol{\hat{\mathcal{E}}} = \boldsymbol{\hat{\epsilon}}_{\theta}^{(t)}}
        $
    \fontsize{8.6}{16} \EndIf
    \State \fontsize{8.6}{16} $\bm{\epsilon}_t \sim \mathcal{N}(0, \bm{I})$
    \State \fontsize{8.6}{16} $\begin{aligned} \mathbf{x}_{t-1} &= \sqrt{\alpha_{t-1}}\left(\frac{\mathbf{x}_t - \sqrt{1 - \alpha_t}\bm{\mathcal{E}}^{'}}{\sqrt{\alpha_t}}\right) \\ &+ \sqrt{1 - \alpha_{t-1} - \left| \bm{\Sigma}_t \right|^{1/d}}\bm{\mathcal{E}}^{'} + \bm{\Sigma}_t^{1/2} \cdot \bm{\epsilon}_t
    \end{aligned}$
    \EndFor
    \State $\mathbf{x}_0^i = \mathbf{x}_0$
\EndFor
\State \Return $\{\mathbf{x}_0^i\}_{i=1}^N$
\end{algorithmic}
\label{alg:dddpm}
\end{algorithm}
\section{Experiments}

\subsection{Experiments Settings}

\textbf{Dataset and Metrics.} We evaluated proposed D$^2$-DPM using LDM \cite{rombach2022high} across three standard datasets: ImageNet, LSUN-Bedrooms, and LSUN-Churches \cite{yu2015lsun}, each with a resolution of 256$\times$256. To quantify generation performance, we employ metrics such as Frechet Inception Distance (FID), Sliding Fréchet Inception Distance (sFID), Inception Score (IS), precision, and recall for comprehensive evaluation. For each evaluation, we generate 50,000 samples and calculate these metrics using the OpenAI's evaluator \cite{dhariwal2021diffusion}, with BOPs (Bit Operations) as the efficiency metric.

\noindent\textbf{LDM settings.} We primarily focus on the generative sampler parameters in LDM: classifier-free guidance $scale$, sampling $step$ and variance schedule $\eta$. Since LDM employs the DDIM sampler, it degrades to an ODE-based sampler with zero stochasticity capacity when $\eta = 0$, becomes an SDE-based DDPM sampler with inherent stochasticity capacity when $\eta = 1$. Therefore, we simulate stochasticity capacity changes by adjusting the $scale$. In class-conditional generation, we set four parameter configurations: \{$scale$ = 3.0, $\eta$ = 0.0$|$1.0, $steps$ = 20\} and \{$scale$ = 1.5, $\eta$ = 0.0$|$1.0, $steps$ = 250\}. For unconditional generation, we set two parameter configurations: \{$\eta$ = 0.0$|$1.0, $steps$ = 200\}.

\noindent\textbf{Quantization Settings.} We employ BRECQ \cite{li2021brecq} as the PTQ baseline for extensive comparative experiments and implement an LDM-compatible version of Qrop \cite{wei2022qdrop}. To ensure comparability, we keep all settings aligned with PTQD, specifically: 1) using Adaround \cite{nagel2020up} as the weight quantizer; and 2) fixing the first and last layers to 8 bits, while quantizing other layers to the target bit-width. For calibration, we collect the diffusion model's inputs at each sampling timestep as the calibration set. Notationally, WxAy indicates that weights and activations are quantized to x and y bits. In all experiments, we adopt two quantization configurations: W8A8 and W4A8. 
\begin{table*}[h]
\centering
\small
\resizebox{\textwidth}{!}{%
\begin{tabular}{cccccccccc}
\toprule
\textbf{Model} & \textbf{Method} & \textbf{Bits (W/A)} & \textbf{Size (MB)} & \textbf{BOPs (T)} & \textbf{IS $\uparrow$} & \textbf{FID $\downarrow$} & \textbf{sFID $\downarrow$} & \textbf{Precision $\uparrow$} & \textbf{Recall $\uparrow$} \\ 
\midrule
\multirow{14}{*}{\begin{tabular}[c]{@{}c@{}}LDM-4\\$\left(\eta = 0.0\right)$\end{tabular}}
\multirow{3}{*}{} & {FP} & 32/32 & 1742.72 & 102.20 & 366.03 & 11.13 & 7.834 & 93.93$\%$ & 27.98$\%$ \\ 
\cmidrule(lr){2-3} \cmidrule(lr){4-5} \cmidrule(lr){6-10}
& PTQ4DM & 8/8 & 436.79 & 8.76 & 324.21 & 9.37 & 9.87 & 87.15$\%$ & 31.77$\%$ \\ 
& {Q-diffusion} & 8/8 & 436.79 & 8.76 & 327.16 & 8.72 & 10.46 & 86.91$\%$ & 33.26$\%$ \\ 
& PTQD & 8/8 & 436.79 & 8.76 & 324.64 & 8.46 & 10.12 & 87.68$\%$ & 34.64$\%$ \\ 
& Ours$_{S\text{-}D^2}$ & 8/8 & 436.79 & 8.76 & 332.55 & \textbf{8.11} & 8.02 & 87.35$\%$ & 36.55$\%$ \\ 
& Ours$_{D\text{-}D^2}$ & 8/8 & 436.79 & 8.76 & \textbf{333.89} & 8.12 & \textbf{7.92} & \textbf{88.56$\%$} & \textbf{36.69}$\%$ \\ 
\cmidrule(lr){2-3} \cmidrule(lr){4-5} \cmidrule(lr){6-10}
& PTQ4DM & 4/8 & 219.12 & 4.38 & 336.28 & 10.45 & 13.94 & 90.61$\%$ & 28.63$\%$ \\ 
& Q-diffusion & 4/8 & 219.12 & 4.38 & 347.52 & 11.13 & 9.07 & 90.89$\%$ & 29.39$\%$ \\ 
& PTQD & 4/8 & 219.12 & 4.38 & 355.10 & 10.41 & 8.45 & 92.13$\%$ & 27.54$\%$ \\ 
& Ours$_{S\text{-}D^2}$ & 4/8 & 219.12 & 4.38 & \textbf{358.14} & 9.75 & \textbf{6.60} & \textbf{92.25}$\%$ & \textbf{30.21}$\%$ \\ 
& Ours$_{D\text{-}D^2}$ & 4/8 & 219.12 & 4.38 & 357.57 & \textbf{9.71} & 6.65 & 92.22$\%$ & 30.21$\%$ \\ 
\midrule
\multirow{14}{*}{\begin{tabular}[c]{@{}c@{}}LDM-4\\$\left(\eta = 1.0\right)$\end{tabular}}
\multirow{3}{*}{} & FP & 32/32 & 1742.72 & 102.20 & 361.84 & 13.83 & 20.56 & 92.22$\%$ & 19.58$\%$ \\ 
\cmidrule(lr){2-3} \cmidrule(lr){4-5} \cmidrule(lr){6-10}
& PTQ4DM & 8/8 & 436.79 & 8.76 & 332.18 & 12.24 & 18.63 & 87.21$\%$ & 23.60$\%$ \\ 
& Q-diffusion & 8/8 & 436.79 & 8.76 & 335.61 & 11.07 & 16.15 & 88.50$\%$ & 24.93$\%$ \\ 
& PTQD & 8/8 & 436.79 & 8.76 & 335.70 & 10.86 & 15.02 & 88.44$\%$ & 25.24$\%$\\ 
& Ours$_{S\text{-}D^2}$ & 8/8 & 436.79 & 8.76 & 342.71 & \textbf{10.57} & 14.81 & 88.58$\%$ & 26.02$\%$ \\ 
&  Ours$_{D\text{-}D^2}$ & 8/8 & 436.79 & 8.76 & \textbf{343.68} & 10.58 & \textbf{14.72} & \textbf{88.90$\%$} & \textbf{26.03}$\%$ \\ 
\cmidrule(lr){2-3} \cmidrule(lr){4-5} \cmidrule(lr){6-10}
& PTQ4DM & 4/8 & 219.12 & 4.38 & 340.10 & 13.68 & 22.05 & 89.50$\%$ & 20.01$\%$ \\ 
& Q-diffusion & 4/8 & 219.12 & 4.38 & 349.89 & 14.22 & 20.17 & 89.93$\%$ & 20.57$\%$ \\ 
& PTQD & 4/8 & 219.12 & 4.38 & 353.57 & 13.15 & 17.41 & 91.09$\%$ & 20.42$\%$ \\ 
& Ours$_{S\text{-}D^2}$ & 4/8 & 219.12 & 4.38 & 355.20 & \textbf{12.60} & 15.80 & 91.26$\%$ & 20.94$\%$ \\  
&  Ours$_{D\text{-}D^2}$ & 4/8 & 219.12 & 4.38 & \textbf{356.39} & 12.65 & \textbf{15.47} & \textbf{91.44$\%$} & \textbf{21.36$\%$} \\  
\bottomrule
\end{tabular}%
}
\caption{Performance comparison of class-conditioned generation on ImageNet 256$\times$256 using LDM-4 (scale=3.0, step=20).}
\label{table:1}
\end{table*}
\subsection{Class-conditional Generation}
We first compare the proposed D$^2$-DPM with other works on class-conditional generation tasks. We conduct experiments using LDM-4 on ImageNet 256$\times$256. The results for the configuration \{$scale$ = 3.0, $\eta$ = 0.0$|$1.0, $steps$ = 20\} are presented in Table \ref{table:1}. Regarding efficiency, W8A8 and W4A8 quantization achieve volume compression ratios of 3.99x and 7.95x, while reducing BOPs by 11.67x and 23.33x. In terms of generation quality, S-D$^2$ and D-D$^2$ demonstrates superior performance across various sampling stochasticity capacities and quantization bit-width settings. Specifically, S-D$^2$ outperforms other works across all metrics, with FID scores in the best-case scenario up to 0.66 lower than PTQD, 1.26 lower than PTQ4DM, and 3.24 lower than the full-precision model. This initially demonstrates that the proposed quantization noise model precisely captures quantization noise during inference. Subsequently, S-D$^2$ effectively restores the distribution by implicitly correcting the standard deviation while correcting the mean. Additionally, its inherent stochasticity accumulates positive effects over prolonged iterations, steering the data toward a more optimal distribution. Moreover, the superior metrics of D-D$^2$ reaffirm the efficacy of noise modeling and distribution correction. When the stochastic capacity $g(t)$ is sufficiently large ($\eta = 1.0$), FID and sFID decrease by 1.06 and 5.47 on average compared to the full-precision model. This indicates that additional standard deviation is effectively utilized by the stochastic term without causing detrimental variance overflow. However, data shows that even when $g(t)$ is too minimal to absorb additional standard deviation, the performance remains superior. We attribute this to the limited additional variance from quantization noise effectively compensating for the stochastic term, creating a superior Langevin SDE over the original SDE (ODE), as discussed in preliminary work. This constructs a larger error buffer, smoothing out the sharp noise introduced at each step. Therefore, we suggest using the D-D$^2$ optimized low-stochasticity-capacity sampler when the quantization bit-width is not too low, meaning the variance from quantization noise is limited, as it can effectively leverage the beneficial components of the noise.

To further validate the performance of our D$^2$-DPM, we conduct two sets of high-density step generation experiments under the conditions \{$scale$ = 1.5, $\eta$ = 0.0$|$1.0, $steps$=250\}, with the results shown in Table \ref{table:2}. Evidently, our metrics consistently surpass PTQD, which demonstrated strong competitive advantages in earlier experiments.

\begin{table*}[ht!]
\centering
\small
\resizebox{\textwidth}{!}{%
\begin{tabular}{c c c c c c c c c c}
\toprule
\textbf{Model} & \textbf{Method} & \textbf{Bits (W/A)} & \textbf{Size (MB)} & \textbf{BOPs (T)} & \textbf{IS $\uparrow$} & \textbf{FID $\downarrow$} & \textbf{sFID 
 $\downarrow$} & \textbf{Precision $\uparrow$} & \textbf{Recall $\uparrow$} \\ 
\midrule
\multirow{4}{*}{\begin{tabular}[c]{@{}c@{}}LDM-4 \\$\left(\eta = 0.0\right)$\end{tabular}}
& FP & 32/32 & 1742.72 & 102.20 & 213.74 & 3.32 & 5.23 & 83.04$\%$ & 53.31$\%$ \\ 
\cmidrule(lr){2-4} \cmidrule(lr){5-6} \cmidrule(lr){7-10}
& PTQD & 4/8 & 219.12 & 4.38 & 162.77 & 6.46 & 10.14 & 73.88$\%$ & 58.10$\%$ \\ 
& Ours$_{D\text{-}D^2}$ & 4/8 & 219.12 & 4.38 & \textbf{169.03} & \textbf{5.56} & \textbf{9.45} & \textbf{75.20$\%$} & \textbf{58.14$\%$} \\ 
\midrule
\multirow{4}{*}{\begin{tabular}[c]{@{}c@{}}LDM-4\\$\left(\eta = 1.0\right)$\end{tabular}}
& FP & 32/32 & 1742.72 & 102.20 & 250.97 & 3.54 & 5.07 & 87.10$\%$ & 49.12$\%$ \\ 
\cmidrule(lr){2-4} \cmidrule(lr){5-6} \cmidrule(lr){7-10}
& PTQD & 4/8 & 219.12 & 4.38 & 153.01 & 7.90 & 7.87 & 71.75$\%$ & 55.15$\%$ \\ 
& Ours$_{S\text{-}D^2}$ & 4/8 & 219.12 & 4.38 & \textbf{171.49} & \textbf{6.91} & \textbf{7.49} & \textbf{72.82$\%$} & \textbf{55.64$\%$} \\  
\bottomrule
\end{tabular}%
}
\caption{Performance comparisons of class-conditional generation on ImageNet256$\times$256 using LDM-4 (scale=1.5, step=250).}
\label{table:2}
\end{table*}
\subsection{Unconditional Generation}
\begin{table}[t]
\centering
\small
\begin{tabular}{cccccc}
\toprule
\multicolumn{6}{c}{LDM-4 (steps=200, $\eta=1.0$)} \\
\midrule
\textbf{Method} & \textbf{W/A} & \textbf{FID $\downarrow$} & \textbf{sFID $\downarrow$} & \textbf{Prec. $\uparrow$} & \textbf{Rec. $\uparrow$}  \\  
\midrule
FP & 32/32 & 3.03 & 7.03 & 64.65$\%$ & 47.60$\%$\\ 
\midrule
PTQD & 8/8 & 9.16 & 12.94 & 51.99$\%$ & 44.32$\%$ \\ 
Ours$_{S\text{-}D^2}$ & 8/8 & 7.69 & 12.61 & 54.81$\%$ & 45.03$\%$\\ 
Ours$_{D\text{-}D^2}$ & 8/8 & \textbf{7.55} & \textbf{12.56} & \textbf{55.60}$\%$ & \textbf{45.80}$\%$\\ 
\cmidrule(lr){1-2} \cmidrule(lr){3-6} 
PTQD & 4/8 & 12.57 & 16.04 & 51.31$\%$ & 42.40$\%$\\ 
Ours$_{S\text{-}D^2}$ & 4/8 & 11.26 & 15.60 & \textbf{51.45$\%$} & 43.66$\%$ \\
Ours$_{D\text{-}D^2}$ & 4/8 & \textbf{10.72} & \textbf{15.23} & 51.44$\%$ & \textbf{43.90$\%$}
\\  
\bottomrule
\end{tabular}%
\caption{\justifying Performance comparisons of unconditional image generation on LSUN-Bedroom 256$\times$256.}
\label{table:3}
\end{table}
We evaluate D$^2$-DPM on unconditional generation tasks, employing LDM-4 and LDM-8 models across the LSUN-Bedroom and LSUN-Church datasets, respectively. Table \ref{table:3} and \ref{table:4} show that our approach narrows the gap with full-precision model. Specifically, on LSUN-Bedroom dataset, S-D$^2$ reduces FID and sFID by 1.39 and 0.39 on average, compared to PTQD. Similarly, D-D$^2$ reduces FID and sFID by 1.73 and 0.60 on average. On LSUN-Church dataset, S-D$^2$ reduces the average FID and sFID by 1.14 and 0.37. In parallel, D-D$^2$ also achieves significant reductions, lowering FID and sFID by an average of 1.13 and 0.39. This demonstrates that our precise quantization noise modeling, along with the decoupled mean and standard deviation corrections in D$^2$-DPM, more effectively restores the distribution. Finally, we observe a phenomenon consistent with previous findings: even when $g(t)$ is minimal, D-D$^2$ still shows superior performance, even partially surpassing S-D$^2$. This confirms the effectiveness of its deterministic mean correction and further validates our earlier hypothesis: the additional variance from quantization noise effectively compensates for the stochastic term, thereby implicitly transforming the sampling equation into the original ODE with an enhanced Langevin diffusion term. The ODE aligns the marginal distribution $p_t(x)$, while the improved Langevin diffusion term better buffers against sharp noise from the quantized diffusion model during each iteration of noise estimation.
\begin{table}[h]
\centering
\footnotesize
\begin{tabular}{cccccc}
\toprule
\multicolumn{6}{c}{LDM-8 (steps = 200, $\eta = 0.0$)} \\
\midrule
\textbf{Method} & \textbf{W/A} & \textbf{FID $\downarrow$} & \textbf{sFID $\downarrow$} & \textbf{Prec. $\uparrow$} & \textbf{Rec. $\uparrow$}  \\ 
\midrule
FP & 32/32 & 4.17 & 12.91 & 66.00$\%$ & 51.46$\%$ \\ 
\midrule
PTQD & 8/8 & 8.31 & 12.97 & 56.57$\%$ & 54.15$\%$ \\ 
Ours$_{S\text{-}D^2}$ & 8/8 & \textbf{7.82} & 12.52 & 56.75$\%$ & 54.45$\%$ \\ 
Ours$_{D\text{-}D^2}$ & 8/8 & 7.83 & \textbf{12.51} & \textbf{56.86}$\%$ & \textbf{54.54}$\%$  \\ 
\cmidrule(lr){1-2} \cmidrule(lr){3-6} 
PTQD & 4/8 & 12.96 & 15.42 & 50.23$\%$ & 52.80$\%$ \\ 
Ours$_{S\text{-}D^2}$ & 4/8 & \textbf{11.18} & 15.14 & \textbf{52.27$\%$} & \textbf{53.78$\%$} \\ 
Ours$_{D\text{-}D^2}$ & 4/8 & 11.18 & \textbf{15.11} & 52.15$\%$ & 53.68$\%$  \\ 
\bottomrule
\end{tabular}%
\caption{\justifying Performance comparisons of unconditional image generation on LSUN-Church 256$\times$256.}
\label{table:4}
\end{table}

\begin{table}[h]
\centering
\footnotesize
\begin{tabular}{cccccc}
\toprule
\textbf{Method} & \textbf{W/A} & \textbf{IS $\uparrow$} & \textbf{FID $\downarrow$} & \textbf{sFID $\downarrow$} \\ 
\midrule
FP & 32/32 & 250.97 & 3.54 & 5.07 \\ 
\midrule
PTQD & 4/8 & 153.01 & 7.90 & 7.87 \\ 
+ SJC (S-D$^2$) & 4/8 & 171.49 & 6.91 & 7.49 \\ 
+ DMC & 4/8 & 159.30 & 7.14 & 7.67  \\ 
+ DMC + DVC (D-D$^2$) & 4/8 & 172.13 & 6.81 & 7.42 \\ 
\bottomrule
\end{tabular}%
\caption{\justifying Ablation study of denoising components using LDM-4 (scale=1.5,$\eta$=1.0,step=250) on ImageNet 256$\times$256.}
\label{table:5}
\end{table}

\subsection{Ablation Study}
As shown in Table~\ref{table:5}, we perform ablation studies on the denoising components of dual denoising mechanisms, S-D$^2$ and D-D$^2$. Stochastic Joint Correction (SJC), which implicitly corrects the variance while correcting the mean using estimated noise, corresponds to S-D$^2$, while Deterministic Mean Correction (DMC) and Deterministic Variance Correction (DVC) are the key components of D-D$^2$. By applying SJC, we achieve FID and sFID reductions of 0.99 and 0.38 compared to PTQD, showing that S-D$^2$ successfully performs joint correction for more accurate distribution restoration. In D-D$^2$, the use of DMC alone reduces FID and sFID by 0.76 and 0.20, respectively, showing that we accurately estimate the conditional mean of quantization nois through joint distribution. Building on this, applying DVC further reduces FID and sFID by 0.33 and 0.25, indicating that the additional variance is also effectively absorbed by the stochastic term. The above experiments show that our D$^2$-DPM more effectively mitigates the adverse effects of quantization noise and more precisely corrects distributions.

\section{Conclusion and Future Work}
In this paper, we propose a dual denoising paradigm to eliminate the residual quantization noise in quantized diffusion models. We first establish the joint distribution of quantized outputs and noise, allowing us to instantiate the conditional distribution of quantization noise during inference. We then design two variants, S-D$^2$ and D-D$^2$, to decouple and correct the mean and standard deviation shifts introduced by quantization noise. Extensive experiments demonstrate that our approach effectively corrects the distribution, achieving high-fidelity quantization of diffusion models.

In essence, this method provides technical support for high-fidelity, efficient compression of diffusion models aimed at reducing carbon emissions, and is therefore not limited by task type. It can be extended to various domains, including video generation, text modeling, and molecular design. Future work will focus on expanding this paradigm across multiple domains and pursuing a unified framework.

\section{Acknowledgments}
This work is supported by Zhejiang Province High-Level Talents Special Support Program "Leading Talent of Technological Innovation of Ten-Thousands Talents Program" (No. 2022R52046) and Alibaba-Zhejiang University Joint Research Institute of Frontier Technologies.

\bibliography{aaai25}

\end{document}